\documentclass[lettersize,journal]{IEEEtran}
\usepackage{amsmath,amsfonts}
\usepackage{algorithm}
\usepackage{array}
\usepackage[caption=false,font=normalsize,labelfont=sf,textfont=sf]{subfig}
\usepackage{textcomp}
\usepackage{stfloats}
\usepackage{url}
\usepackage{verbatim}
\usepackage{graphicx}
\usepackage{cite}

\usepackage{makecell}
\usepackage{utfsym}
\usepackage{algpseudocode}
\usepackage{algorithmicx}
\usepackage{multirow}
\usepackage{colortbl}
\usepackage{amssymb}
\usepackage{url}

\begin{document}

\title{Exploring Fine-Grained Representation and Recomposition for Cloth-Changing Person Re-Identification}

\author{
Qizao Wang, Xuelin Qian, Bin Li, Xiangyang Xue, and Yanwei Fu
\thanks{Qizao Wang, Bin Li and Xiangyang Xue are with the School of Computer Science, and Shanghai Key Lab of Intelligent Information Processing, Fudan University, Shanghai 200437, China. Email: qzwang22@m.fudan.edu.cn, \{libin, xyxue\}@fudan.edu.cn.}
\thanks{Xuelin Qian is with the School of Automation, Northwestern Polytechnical University, Xi'an 710021, China. Email: xlqian@nwpu.edu.cn. 
Xuelin Qian is the corresponding author.}
\thanks{Yanwei Fu is with the School of Data Science, and Fudan ISTBI—ZJNU Algorithm Centre for Brain-inspired Intelligence, Fudan University, Shanghai 200437, China. Email: yanweifu@fudan.edu.cn.}
}

\markboth{IEEE TRANSACTIONS ON INFORMATION FORENSICS AND SECURITY}{}


\maketitle

\begin{abstract}
Cloth-changing person Re-IDentification (Re-ID) is a particularly challenging task, suffering from two limitations of inferior discriminative features and limited training samples. Existing methods mainly leverage auxiliary information to facilitate identity-relevant feature learning, including soft-biometrics features of shapes or gaits, and additional labels of clothing. However, this information may be unavailable in real-world applications. In this paper, we propose a novel FIne-grained Representation and Recomposition (FIRe$^{2}$) framework to tackle both limitations without any auxiliary annotation or data. Specifically, we first design a Fine-grained Feature Mining (FFM) module to separately cluster images of each person. Images with similar so-called fine-grained attributes (\textit{e.g.}, clothes and viewpoints) are encouraged to cluster together. An attribute-aware classification loss is introduced to perform fine-grained learning based on cluster labels, which are not shared among different people, promoting the model to learn identity-relevant features. Furthermore, to take full advantage of fine-grained attributes, we present a Fine-grained Attribute Recomposition (FAR) module by recomposing image features with different attributes in the latent space. It significantly enhances robust feature learning. Extensive experiments demonstrate that FIRe$^{2}$ can achieve state-of-the-art performance on five widely-used cloth-changing person Re-ID benchmarks. 
The code is available at \url{https://github.com/QizaoWang/FIRe-CCReID}.
\end{abstract}


\section{Introduction}
\label{sec:intro}

\IEEEPARstart{P}{erson} Re-Identification (Re-ID) targets to identify the same person across different cameras, which has great potential in video surveillance applications. In the past decades, plenty of efforts~\cite{li2018harmonious,sun2018beyond,hou2019interaction,qian2019leader,wang2018learning,wang2023rethinking} have been made to promote development in this field and achieved excellent performance. However, they mainly focus on short-term scenarios with an impractical assumption that the same person would wear the same clothes. As a result, a new task of cloth-changing person Re-ID is recently derived to pursue a more robust model in long-term scenarios.
Some of the existing approaches~\cite{xu2021adversarial,eom2021disentangled} introduce generative models to synthesize images of the same person with various clothes, so as to explore cloth-irrelevant features. Other researchers \cite{qian2020long,yang2019person,wang2022co,jin2022cloth,hong2021fine,chen2021learning} concentrate on drawing support from auxiliary modalities, such as keypoints, contours, gaits, and 3D shapes. However, these two solutions require either huge amounts of additional training data for generative models or well-trained off-the-shelf tools for modality extraction, which is uncontrollable and burdensome.

\begin{figure}[t]
    \centering
    \includegraphics[width=0.94\linewidth]{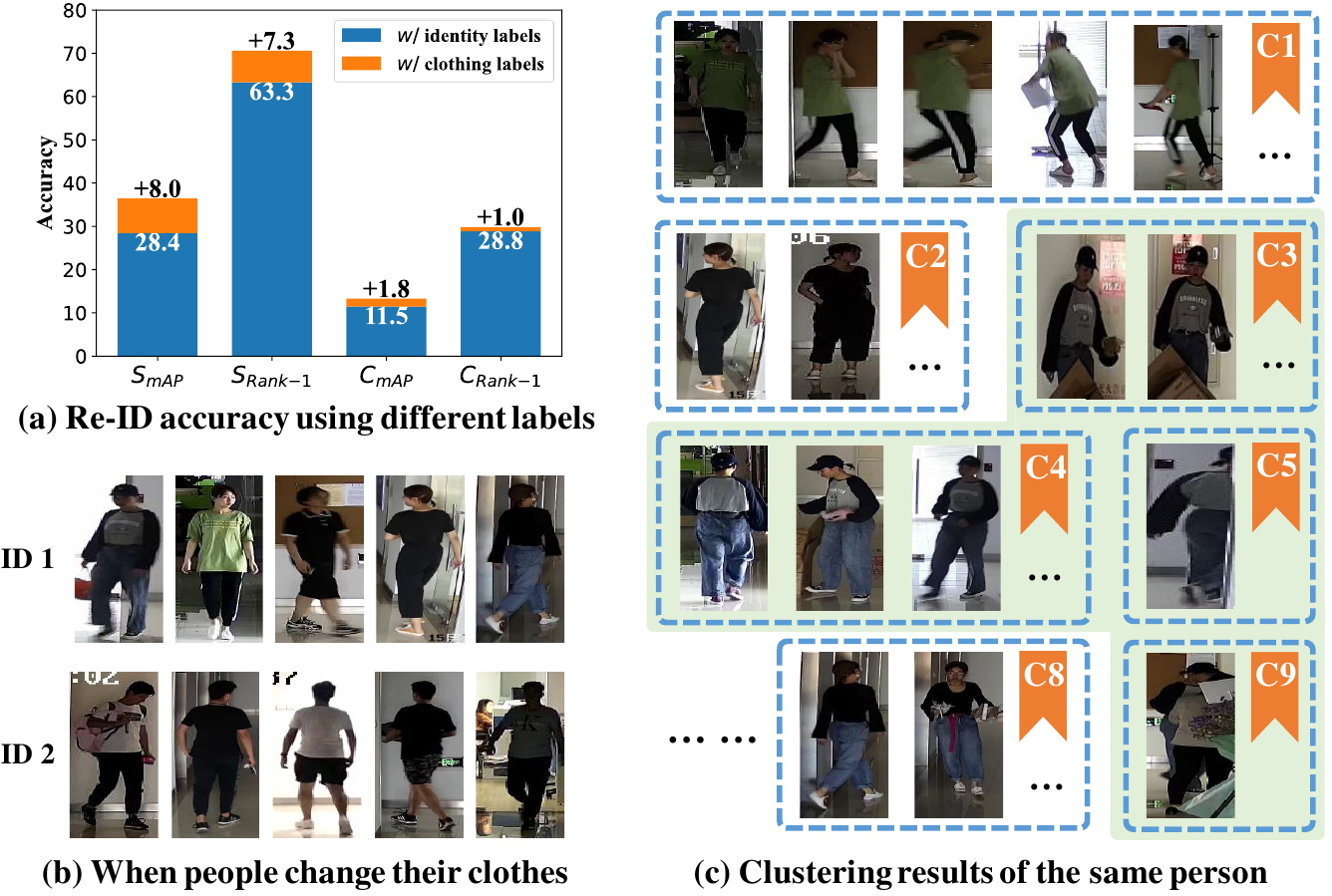}
    \vspace{-0.05in}
    \caption{\textbf{Pilot studies to support our motivation.}
    (a) We train a vanilla ResNet-50~\cite{resnet} with identity labels or clothing labels on the LTCC dataset. ``S'' and ``C'' denote the standard and the cloth-changing settings, respectively. Clothing labels bring more improvement, especially in the standard setting. However, it is difficult to define and annotate labels of various clothing styles. (b) When pedestrians change clothes, more fine-grained clues than clothing are needed to determine their identity. (c) We show clustering results of images from the same person. Shared fine-grained attributes (\textit{e.g.}, clothes, viewpoint, and occlusion) can be easily found in each cluster (C1 $\sim$ C9). Images in the green region, while having the same ground-truth clothing labels, can be further divided according to other different fine-grained attributes.
    }
    \label{fig:intro}
    \vspace{-0.1in}
\end{figure}

Recently, Gu~\textit{et al.}~\cite{gu2022clothes} promote this task by utilizing more lightweight clothing label information, which catches our attention in particular. Clothing labels are a subcategory of identities, containing finer-grained information\footnote{Clothing labels in existing person Re-ID datasets are annotated based on identities, \textit{i.e.}, different people won't have the same clothing label.\\}.
A natural question thus lingers in our minds: \textit{is fine-grained information conducive to identity-relevant feature learning?} To answer it, we conduct a pilot study in Fig.~\ref{fig:intro} (a). The positive answer suggests that using just one of fine-grained information, \textit{i.e.}, clothing labels, is already instructive for person Re-ID, especially in the standard setting. One intuitive explanation is that learning fine-grained information can encourage the model to focus on more details of person images.
However, we notice several limitations to following this pipeline. \textbf{(1)} Clothing labels require statistical information of the whole dataset, which is unavailable for practical applications. Besides, it is ambiguous for annotators to define labels of various combinations of clothes. 
\textbf{(2)} When pedestrians change clothes, more fine-grained clues than clothing are needed to determine their identity, as shown in Fig.~\ref{fig:intro} (b). Nevertheless, existing Re-ID datasets mostly lack labels of fine-grained information, and it is cumbersome and impractical to manually annotate each sample. 
\textbf{(3)} Due to the difficulty of data collection and annotation, the amount of fine-grained information often varies greatly from person to person, and the total number is insufficient, leading to sub-optimal benefits in the cloth-changing setting, as revealed in Fig.~\ref{fig:intro} (a).

To this end, we propose a novel FIne-grained Representation and Recomposition (FIRe$^{2}$) framework to instantiate the aforesaid pipeline by addressing the above limitations. 
Our key idea is to explore the fine-grained information (attributes) of each person through clustering, and then facilitate the model learning with fine-grained representation learning. \textit{``Fine-grained attributes'' here are a subcategory of identities, and refer to clothes, viewpoints and anything we get from clustering.}
Concretely, we first propose a Fine-grained Feature Mining (FFM) module. Considering that labels of fine-trained attributes are difficult to obtain, samples with the same identity are clustered according to their features, and then assigned to different fine-grained pseudo labels. As depicted in Fig.~\ref{fig:intro}~(c), shared fine-grained attributes are discovered from images in every cluster. 
Then, an attribute-aware classification loss is introduced to facilitate fine-grained learning by classifying images close to the corresponding fine-grained pseudo labels. Such a simple strategy can not only discard the requirement for extra labels (\textit{e.g.}, clothing), but also mine rich fine-grained attributes for discriminative identity-relevant feature learning. 

To tackle the problem of limited data, we present a Fine-grained Attribute Recomposition (FAR) module, serving as data augmentation, to take full advantage of learned fine-grained attributes. Specifically, we adopt instance normalization~\cite{ulyanov2016instance} to disentangle the original attribute of the input image, and then recompose different attributes between images in the same batch. Furthermore, we separately recompose the upper and lower body attributes of the input image, to enrich the representation of the same person with various attributes. It significantly encourages models to learn robust identity-relevant features from these representations. 

Overall, our proposed framework not only leverages fine-grained learning to mine discriminative features from person images, but also realizes feature augmentation using fine-grained attributes. More importantly, it does not rely on any auxiliary information, like clothing labels or other modalities, and can freely discard the proposed two modules during inference, showing great potential for real-world applications. 

\noindent \textbf{Contributions.} We summarize the key contributions as follows,
\textbf{(1)} We propose a novel framework FIRe$^{2}$ for cloth-changing person Re-ID. It only requires RGB images as input, but can extract identity-relevant and cloth-irrelevant features.
\textbf{(2)} We design a Fine-grained Feature Mining (FFM) module to acquire fine-grained attributes via clustering. An attribute-aware classification loss is further introduced to facilitate fine-grained representation learning.
\textbf{(3)}  We present a Fine-grained Attribute Recomposition (FAR) module to effectively augment the features of each image with various attributes, which significantly promotes the learning of robust features.
\textbf{(4)} FIRe$^{2}$ achieves state-of-the-art performance on five popular cloth-changing benchmarks without bells and whistles. Thorough ablation studies and in-depth discussions are provided to show the superiority of our proposed modules.

\section{Related work}
\subsection{Conventional Person Re-Identification}
Person Re-ID has received increasing attention due to its wide range of surveillance applications in the real world. Existing deep models are principally studied under short-term scenarios, where the appearance of the same person is more or less consistent. Great efforts have been made to deal with multiple challenges, including occlusion~\cite{yan2021occluded,wang2022feature}, illumination~\cite{huang2019illumination}, viewpoint~\cite{sun2019dissecting,zhu2020aware}, cross-modalities ~\cite{wu2021discover,lin2022learning} and generalization~\cite{hu2014cross,liao2020interpretable}. 
On the other hand, due to the difficulty of annotating identities, many researchers turn to train person Re-ID models without using any labels, thus deriving a popular topic of unsupervised person Re-ID~\cite{ge2020self,dai2022cluster,zhang2022implicit}. Classical approaches to solving this task are principally based on clustering~\cite{lin2019bottom,ge2020self,dai2022cluster} or memory~\cite{zhong2019invariance,zhong2020learning} mechanism. For clustering-based methods, different pedestrians may inevitably be clustered due to similar visual features. Therefore, BUC~\cite{lin2019bottom} proposes a bottom-up clustering approach to gradually exploit similarity from diverse unlabeled images; SpCL~\cite{ge2020self} employs self-paced learning to gradually generate more reliable clusters.
Differently, we cluster samples with the same identity to mine fine-grained attributes, thereby achieving effective feature enhancement by recomposing fine-grained attributes of different images.

Person attributes are semantic high-level information that is robust to visual appearance variations and contain information that is highly relevant to a person's identity~\cite{schumann2017person}. Attributes as auxiliary information have been used in many person Re-ID works~\cite{li2015multiattr,peng2016joint,schumann2017person,lin2019improving}.
However, they require defining and annotating attributes manually. Traditionally, attributes are defined as age, gender, clothing style, and so on. They are circumscribed and difficult to scale in complicated real-world scenarios, and labeling them is time-consuming and costly. Differently, in this paper, we advocate using the feature similarity of images themselves to efficiently and adaptively mine fine-grained attributes without any supervision or auxiliary dependence.

\subsection{Cloth-Changing Person Re-Identification} 
To break the strong correlation between appearance and identity representation, researchers concentrate on the study of cloth-changing person Re-ID, where pedestrians with various suits are captured in a long-term scenario. Clothing changes not only bring dramatic visual interference, but also lead to difficulties in data acquisition and labeling. 
Xu~\textit{et al.}~\cite{xu2021adversarial} and Eom \textit{et al.}~\cite{eom2021disentangled} attempt to use generative models to augment samples by explicitly synthesizing person images with various clothes, so as to learn more robust features against clothing changes. To keep the semantic structure, Pos-Neg~\cite{jia2022complementary} synthesizes in-distribution and out-of-distribution images with the help of human parsing and pose models.
Nevertheless, it is non-trivial to train a powerful generative model and it inevitably introduces the domain gap. To address the essential problem of lack of data, large efforts have been made to collect all kinds of cloth-changing datasets \cite{yang2019person,qian2020long,huang2019beyond,xu2023deepchange,shu2021large}.

Based on these benchmarks, some studies draw support from auxiliary modalities to better leverage soft-biometrics features~\cite{qian2020long,yang2019person,wang2022co,jin2022cloth,hong2021fine,chen2021learning,liu2023dual,guo2023semantic,wang2023exploring}. For instance, SPT+ASE~\cite{yang2019person} utilizes reliable and discriminative curve patterns on the body contour sketch. GI-ReID~\cite{jin2022cloth} learns cloth-agnostic representations by leveraging personal unique and cloth-independent gait information. CACL~\cite{wang2023exploring} involves extra 3D annotations and builds 2D-3D correspondences to learn discriminative shape features. 
Other researchers~\cite{gu2022clothes,yang2023good,cui2023dcr,han2023clothing} take advantage of clothing labels to eliminate the negative effects of clothing features. For instance, CAL~\cite{gu2022clothes} adopts adversarial learning to penalize the model's predictive power to clothes. AIM~\cite{yang2023good} adopts a dual-branch model to simulate causal intervention and eliminate clothing bias. 
However, relying on multi-modal information as well as extra labels limits the flexibility and application of these models. In this paper, we propose to leverage fine-grained representation learning to enhance the discriminative power of image features without any requirement for auxiliaries. In addition, the explored intrinsic fine-grained attributes of images can further be recomposed in the latent space, so as to realize effective augmentation. Following this pipeline, models can learn more robust identity-relevant features both from effective learning and big data perspectives.

\section{Methodology}
\label{sec:method}

Figure~\ref{fig:framework} illustrates the schematic of our proposed FIRe$^{2}$. We start by introducing how to explore fine-grained attributes and perform fine-grained learning through our proposed FFM module (in Sec.~\ref{subsec:FFM}). Next, we elaborate on the proposed FAR module to augment features by recomposing attributes between images (see Sec.~\ref{subsec:FAR}). Lastly, we discuss the training and inference procedures of our framework in Sec.~\ref{subsec:train_test}.

\subsection{Fine-Grained Feature Mining \label{subsec:FFM}}

Clothing labels are regarded as one of fine-grained identity-based attributes. Although it has been shown to be beneficial for discriminative feature learning, it is just as difficult to access or annotate as identity labels in real-world applications. Besides, a person's fine-grained attributes contain more than just clothing information. 
The variable factors in person images include clothing, illumination, posture, occlusion, background, \textit{etc}, which naturally affect the image feature distribution. Inspired by clustering-based methods for unsupervised person Re-ID~\cite{lin2019bottom,ge2020self,dai2022cluster}, we advocate using the feature similarity of images themselves to efficiently and adaptively mine different fine-grained attributes without any supervision. Differently, \textit{clustering is performed on all images of each person individually, rather than images of all people.}
Specifically, given a training dataset $\mathcal{D} = \{ x_{i}, y_{i}\}_{i=1}^{N}$ containing totally $N$ images and $N^{p}$ identities, where $x_{i}$ and $y_{i}$ denotes the $i$-th images and its corresponding identity label, we first build a CNN model $\mathcal{G}$ to extract image features $f_{i} = \mathcal{G}\left(x_{i}\right)$. Considering that different identities may involve different numbers of fine-grained attributes, we thus adopt the DBSCAN~\cite{ester1996density} algorithm for clustering, for which we do not need to specify the cluster number.

\begin{figure}[t]
  \centering
  \includegraphics[width=1\linewidth]{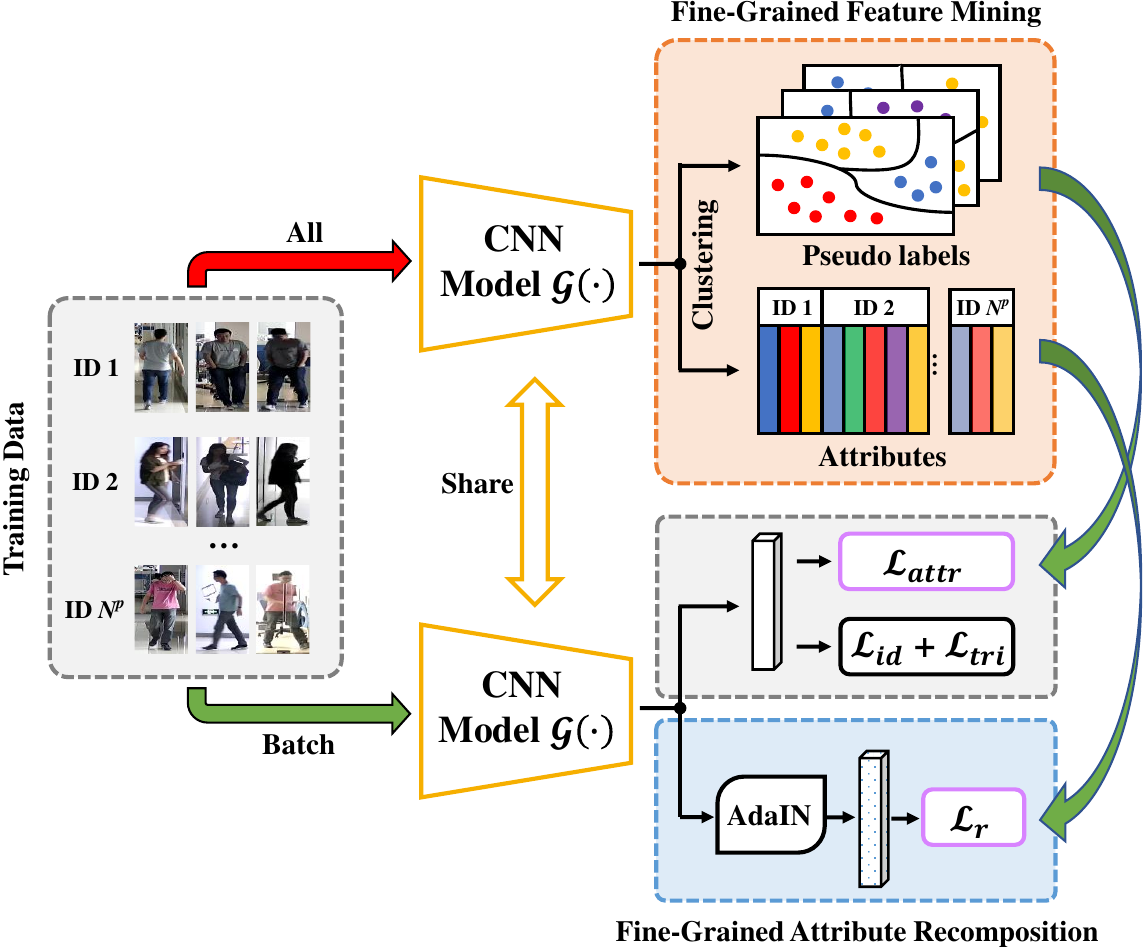}
  \vspace{-0.05in}
  \caption{\textbf{Overview of our proposed method.} 
  We first perform Fine-grained Feature Mining (FFM) to mine fine-grained attributes of all pedestrians. Then, the attribute-aware classification loss $\mathcal{L}_{attr}$ is introduced to encourage fine-grained learning for discriminative features. By taking full advantage of the explored fine-grained attributes, we further present the Fine-grained Attribute Recomposition (FAR) module as an augmentation to recompose features of each image with various attributes, and identity classification loss $\mathcal{L}_{r}$ is applied to facilitate the learning of robust features.}
  \label{fig:framework}
  \vspace{-0.1in}
\end{figure}

As illustrated in the top half of Fig.~\ref{fig:framework}, two important assets can be acquired after clustering. One is fine-grained pseudo labels, denoted by $y^{s}$. Images in the same cluster are assigned to the same pseudo label. However, they are not shared among different people. Denote the number of clusters for each identity as $n^{s}_{j}$, the total number of clusters for $\mathcal{D}$ as $N^{s}= \sum_{j=1}^{N^{p}}n^{s}_{j}$. As a result, we are able to conduct fine-grained learning with $y^{s}$, which will be elaborated on later. 

The other is fine-grained attributes. Inspired by~\cite{ulyanov2016instance}, we utilize instance normalization to embody attributes from image features. Formally, for each image feature $f_{i}$ (before global average/max pooling), we compute the mean and the standard deviation across the spatial dimensions,
\begin{equation}
\small
    \mu_{i}:= \mu\left(f_{i}\right) = \frac{1}{HW}\sum_{h=1}^{H}\sum_{w=1}^{W}f_{i,h,w} 
\label{eq:mu}
\end{equation}
\begin{equation}
\small
    \sigma_{i}:= \sigma\left(f_{i}\right) = \sqrt{\frac{1}{HW}\sum_{h=1}^{H}\sum_{w=1}^{W}\left(f_{i,h,w} - \mu_{i} \right)^{2}}
\label{eq:sigma}
\end{equation}
where $H$ and $W$ indicate the height and width of the feature map. Thereby, we embody and redefine different attributes with a 3-tuple, $\left(\mu_{i}, \sigma_{i}, y^{s}_{i}\right)$. The first two terms describe the distribution of the attribute, and the last term denotes its fine-grained label. In Sec.~\ref{subsec:FAR}, we will show how to use these attributes to enrich the diversity of different image features and effectively enhance the training samples.

\noindent \textbf{Attribute-aware classification loss.}
Once the fine-grained pseudo labels are obtained, we introduce an attribute-aware classification loss to perform fine-grained learning, promoting the model to learn discriminative features from images. Intuitively, we encourage features in the same fine-grained cluster to be pulled closer to each other. On the other hand, the identity supervision is directly related to person Re-ID, so we do not expect to lose this signal while pushing different fine-grained clusters with the same identity away, \textit{i.e.}, the distance between fine-grained clusters of the same person is expected to be smaller than those with different identities. Consequently, we introduce a smooth term $a_{j}$, \textit{w.r.t.} fine-grained pseudo labels, to control the trade-off, which can be expressed as,
\begin{equation}
\small
    \mathcal{L}_{attr} = - \frac{1}{|\mathcal{B}|} \sum\limits_{f \in \mathcal{B}} \sum\limits_{j \in \mathcal{P}} a_{j} \ \log \frac{\exp (f^{\top} \cdot w_{j} / \tau)}{\sum\limits_{k=1}^{N^{s}} \exp ({f^{\top} \cdot w_{k}}  / \tau)}
\label{eq:attr-loss}
\end{equation}
\begin{equation}
\small
    a_{j} = 
    \left\{\begin{matrix} 
    1 - \frac{\left| \mathcal{P} \right| - 1}{\left| \mathcal{P} \right|}\epsilon, \quad j = y^{s} \\  
    \frac{1}{\left| \mathcal{P} \right|} \epsilon, \quad \quad {\rm otherwise}
    \end{matrix}\right.
    , \quad \forall j \in \mathcal{P}
\label{eq:w-loss}
\end{equation} 
where $\mathcal{B}$ means the set of image features in a batch; $\mathcal{P}$ is the index set of clusters with the same identity label; $|\cdot|$ denotes the cardinal number of a set; $\tau$ is the temperature and $\epsilon$ is a hyper-parameter to control the degree of smoothing; $w$ is learnable parameters initialized with fine-grained cluster centers. 

\noindent \textbf{Discussion.} 
By leveraging the proposed fine-grained pseudo labels and attribute-aware classification loss, our model can sensitively capture discriminative clues as well as identity information, to enhance the feature representation learning. Previous work like CAL~\cite{gu2022clothes} takes clothing as disturbing information, and uses clothing labels and adversarial training to disentangle and eliminate clothing information in the image space, respectively.
We point out that it is cumbersome to manually annotate clothing labels, and ambiguous to unify labels of various clothing styles. Differently, clothing information is regarded as one of the instructive fine-grained attributes in our paper. Not only do we not eliminate it, but we encourage the model to extract image features that can distinguish fine-grained information well. More importantly, our proposed FFM module does not require any fine-grained annotation, and is able to mine more fine-grained attributes than clothing, benefiting more identity-relevant feature learning.

\subsection{Fine-Grained Attribute Recomposition \label{subsec:FAR}}

\begin{figure} [t]
  \centering
  \includegraphics[width=0.9\linewidth]{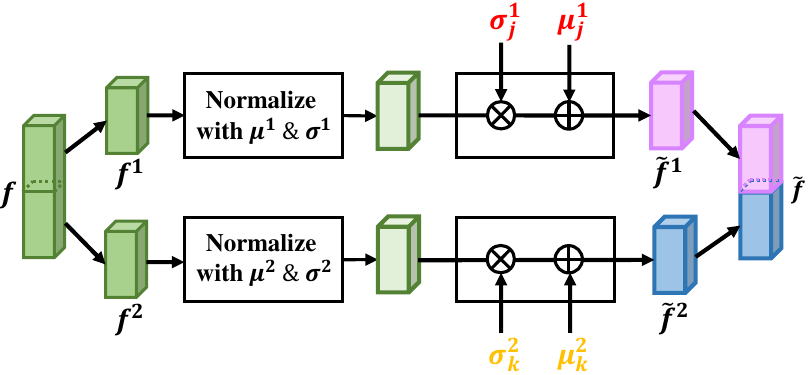}
  \vspace{-0.05in}
  \caption{\textbf{Illustration of fine-grained attribute recomposition.} Taking two parts as an example, it first normalizes the input feature to remove its original attribute in the part level, and then restitutes it with new attributes $\left(\mu_{j}^{1} , \sigma_{j}^{1}\right)$ and $\left(\mu_{k}^{2} , \sigma_{k}^{2}\right)$ from different pedestrians.}
  \label{fig:FAR}
  \vspace{-0.1in}
\end{figure}

Previous data augmentation in person Re-ID is principally based on the pixel space, that is, image enhancement~\cite{zhong2020random} or generation \cite{qian2018pose,eom2021disentangled}. We argue that the quality of generated images is affected by various factors, (\textit{e.g.}, background and illumination), some of which have no direct impact on Re-ID, and even a small deviation may harm the performance (\textit{e.g.}, adversarial samples). Therefore, we propose the Fine-grained Attribute Recomposition (FAR) module. As shown in Fig.~\ref{fig:FAR}, by taking full advantage of fine-grained attributes, it recomposes image features with different attributes in the latent space, so as to effectively enrich the feature representations.

Following \cite{huang2017arbitrary}, we normalize the input image feature $f_{i}$ with instance normalization to remove the original attributes, and then restitute them with attributes with another fine-grained label $\left(\mu_{j}, \sigma_{j}, y^{s}_{j}\right)$, which is formulated as,
\begin{equation}
\small
    \tilde{f}_{i} = \sigma_{j} \cdot \frac{f_{i} - \mu_{i}}{\sigma_{i}} + \mu_{j}
\label{eq:adain}
\end{equation} 

\noindent where $j \in [1, \cdots, N]$ and $y^{s}_{j} \ne y^{s}_{i}$. To train it, we utilize the identity label as supervision and apply cross-entropy loss,
\begin{equation}
\small
    \mathcal{L}_{r} = - \sum\limits_{c=1}^{N^p} q_{c} \log p_{c}, \ 
        q_{c} = 
        \left\{\begin{matrix} 
        1, &y_{i} = c \\  
        0, &{\rm otherwise}
        \end{matrix}\right.
\label{eq:r-loss}
\end{equation}
where $p_{c}$ is the predicted probability of $\tilde{f}_{i}$ for the $c$-th identity class. It is noteworthy that the recomposed features could have fine-grained attributes from either the same person or different people, so as to effectively enrich feature representation. Training with such recomposed features, the model is encouraged to perceive diverse and varying attributes, and learn identity-relevant representations robust to these variations. Experiments in Tab.~\ref{tab:ablation} successfully indicate the superiority of our design.
However, though alternative fine-grained attributes from all images of the dataset are diverse, searching for them is sub-optimal in efficiency. We thus improve it from two aspects.

\noindent \textbf{Recomposition in a batch.} 
As shown in the blue area of Fig.~\ref{fig:framework}, given a set of image features in a batch $\mathcal{B}$, we first retrieve their fine-grained pseudo labels as introduced in Sec.~\ref{subsec:FFM}. For each sample, we then randomly sample an image within $\mathcal{B}$ that does not share the same fine-grained label. Subsequently, we perform Eq.~\ref{eq:adain} to realize the efficient recomposition for feature augmentation.

\noindent \textbf{Recomposition within $P$ parts.} Intuitively, fine-grained attributes vary across the spatial. Taking the clothing attribute as an example, people can wear different jackets or pants, resulting in different styles of clothing. It thus motivates us to perform our proposed attribute recomposition at the local part level. More concretely, we equally divide image features $f_{i}$ horizontally into $P$ parts. Here, fine-grained attributes are also calculated for $P$ parts individually.
For each part $f^{p}_{i}|_{p=1}^{P}$, we perform Eq.~\ref{eq:adain} separately to achieve the augmented counterpart $\tilde{f}^{p}_{i}$. In this way, we recompose each part with one new fine-grained attribute, significantly enriching the feature representation. Finally, they are concatenated together to get $ \tilde{f}_{i} = \left[\tilde{f}^{1}_{i};\tilde{f}^{2}_{i}; \cdots; \tilde{f}^{P}_{i}\right]$.

\algrenewcomment[1]{\hfill\(\triangleright\) \textcolor{gray}{#1}}

\begin{algorithm}[t]
\small
\caption{The training procedure of FIRe$^{2}$}
\label{algorithm}
\begin{algorithmic}
\Require \\
    {\bf Input:}
    CNN model $\mathcal{G}$;
    Training dataset $\mathcal{D} = \{ x_{i}, y_{i}\}_{i=1}^{N}$; \\
    Number of images $N$;
    Number of identities $N^{p}$;\\
    Maximum epoch $\mathcal{T}$; 
    Threshold $t_{0}$ $\left(0 < t_{0} < \mathcal{T}\right)$\\
    {\bf Initialization:} 
    Model weights $\theta_{\mathcal{G}}$; Hyper-parameters $\lambda_{1}, \lambda_{2}, \lambda_{3}, \lambda_{4}$
\end{algorithmic}
\begin{algorithmic}[1]
    \Ensure Optimal weights $\theta^{*}_{\mathcal{G}}$
    
    \State {\bf for} $t \gets 1$ to $\mathcal{T}$ {\bf do}

    \State \quad \textcolor{gray}{$//$ Clustering}
    \State \quad {\bf for} $i \gets 1$ to $N$ {\bf do}\Comment{batch computing}
    
    \State \qquad Extract image features $f_{i} = \mathcal{G}\left(x_{i}\right)$
    \Comment{freeze weights}
    
    \State \quad {\bf end for}
    
    \State \quad {\bf for} $n \gets 1$ to $N^{p}$ {\bf do}\Comment{clustering for each person}
    
    \State \qquad $\mathcal{Q}_{n}=\{f_{i}~|~i \in \left[1,\cdots,N\right], y_{i}=n \}$
    \State \qquad $y^{s} \gets \text{DBSCAN} \left( \mathcal{Q}_{n} \right)$\Comment{fine-grained labels}
    
    \State \quad {\bf end for}

    \State \quad \textcolor{gray}{$//$ Training}
    
    \State \quad {\bf for} $i \gets 1$ to $N$ {\bf do}
    \Comment{batch computing}
    
    \State \qquad Extract features $f_{i} = \mathcal{G}\left(x_{i}\right)$
    \Comment{unfreeze weights}
    
    \State \qquad Define attributes $\left(\mu_{i}, \sigma_{i}, y^{s}_{i}\right)$ via Eqs.~\textcolor{red}{1} and \textcolor{red}{2}

     \State \qquad Get augmented features $\tilde{f}_{i}$ via Eq.~\textcolor{red}{5}

     \State \qquad Compute $\mathcal{L}_{id}$ and $\mathcal{L}_{tri}$ in Eq.~\textcolor{red}{7}
     \Comment{basic Re-ID losses}
     
     \State \qquad {\bf if} $t \le t_{0}$ {\bf then}
     \State \quad \qquad $\mathcal{L} = \lambda_{1} \mathcal{L}_{id}$

     \State \qquad {\bf else}
     
     \State \quad \qquad Compute $\mathcal{L}_{attr}$ via Eqs.~\textcolor{red}{3} and \textcolor{red}{4}
    \State \quad \qquad Compute $\mathcal{L}_{r}$ via Eq.~\textcolor{red}{6}
    \State \quad \qquad $\mathcal{L} = \lambda_{1} \mathcal{L}_{id} + \lambda_{2}\mathcal{L}_{tri} + \lambda_{3}\mathcal{L}_{attr} + \lambda_{4} \mathcal{L}_{r}$

    \State \qquad {\bf end if}
    
    \State \qquad 
    $\theta_{\mathcal{G}} \leftarrow \theta_{\mathcal{G}} - \bigtriangledown_{\theta_{\mathcal{G}}} \mathcal{L}$
    \Comment{backward and update $ \theta_{\mathcal{G}}$}

    \State \quad {\bf end for}
    
    \State {\bf end for}
    \State \Return $\theta^{*}_{\mathcal{G}}= \theta_{\mathcal{G}}$
\end{algorithmic}
\end{algorithm}

\subsection{Training and Inference \label{subsec:train_test}}
As illustrated in Fig.~\ref{fig:framework}, at the beginning of each epoch, we first use the FFM module to explore fine-grained pseudo labels as well as attributes for each sample (red line). Then, we leverage fine-grained learning and the FAR module to iteratively train the model $\mathcal{G}$ with mini-batch data (green line). 

More specifically, in the early training stage, the model tends to learn coarse-grained and easy identity information by distinguishing easy samples (with the same identity and clothes). Our proposed FFM and FAR modules encourage the model to learn fine-grained identity features, so if they are introduced at the early stage, they may lead the model to the local optimum. 
Therefore, before the model has learned decent pedestrian identity representation, we only use the basic identity classification loss $\mathcal{L}_{id}$~\cite{luo2019bag} for supervision. Then, we gradually add other terms, including the widely-used triplet loss $\mathcal{L}_{tri}$~\cite{hermans2017defense}, our proposed attribute-aware classification loss $\mathcal{L}_{attr}$ and cross-entropy loss $\mathcal{L}_{r}$ for attribute-recomposed features, to jointly help the model learn robust fine-grained identity features. The overall loss function is formulated as,
\begin{equation}
\label{eq:overall_loss}
    \mathcal{L} = \lambda_{1} \mathcal{L}_{id} + \lambda_{2}\mathcal{L}_{tri} + \lambda_{3}\mathcal{L}_{attr} + \lambda_{4} \mathcal{L}_{r}
\end{equation}
where $\lambda$ is the coefficient to control the contribution of each term. We empirically set $\lambda_{1}=\lambda_{2}=\lambda_{3}=1$ and pay more attention to $\mathcal{L}_{r}$. 
The training procedure of FIRe$^{2}$ is provided in Alg.~\ref{algorithm}, and 
more discussions can be found in Sec.~\ref{subsec:ablation}.

\noindent \textbf{Inference.}
We use the learned discriminative feature $f$ and compute the cosine distances between two person images as metrics for person Re-ID. During inference, FIRe$^{2}$ can freely discard FFM and FAR modules, thus there is no extra storage or computing overload, showing higher efficiency.

\section{Experiments}
\label{sec:experiment}

\begin{table*}[tp]
\centering
\caption{\textbf{Comparison of our method with state-of-the-art methods on PRCC and LTCC.} 
``sketch'', ``sil.'', ``pose'', ``gray'', and ``3D'' represent the contour sketches, silhouettes, human poses, gray images, and 3D shape information, respectively. ``Cloth-Labels'' indicates whether to use ground-truth clothing labels for training. Methods marked with ``$\ast$'' involve multiple training stages for extra auxiliary networks. ``Standard’' and ``Cloth-Changing'' mean the standard and cloth-changing settings, respectively.
\label{tab:ltcc}}
\vspace{-0.05in}
\setlength{\tabcolsep}{3mm}{
\begin{tabular}{l|c|c|cc|cc|cc|cc}
\Xhline{1pt}
\multicolumn{1}{c|}{\multirow{3}{*}{\textbf{Methods}}} &
  \multirow{3}{*}{\textbf{Modality}} &
  \multirow{3}{*}{\textbf{\begin{tabular}[c]{@{}c@{}}Cloth-\\ Labels\end{tabular}}} &
  \multicolumn{4}{c|}{\textbf{PRCC}} &
  \multicolumn{4}{c}{\textbf{LTCC}} \\ \cline{4-11} 
 &
   &
   &
  \multicolumn{2}{c|}{Cloth-Changing} &
  \multicolumn{2}{c|}{Standard} &
  \multicolumn{2}{c|}{Cloth-Changing} &
  \multicolumn{2}{c}{Standard} \\ \cline{4-11} 
                     &                  &  & Rank-1 & \multicolumn{1}{c|}{mAP} & Rank-1 & mAP  & Rank-1 & \multicolumn{1}{c|}{mAP} & Rank-1 & mAP  \\ \hline \hline
\rowcolor{gray!10} CESD~\cite{qian2020long} & RGB+pose & $\checkmark$ & - & \multicolumn{1}{c|}{-}  & -  & -  & 26.2 & \multicolumn{1}{c|}{12.4} & 71.4   & 34.3 \\
\rowcolor{gray!10} AFD-Net~\cite{xu2021adversarial} & RGB+gray & $\checkmark$ & 42.8 & \multicolumn{1}{c|}{-} & 95.7 & - & - & \multicolumn{1}{c|}{-}    & -      & -    \\
\rowcolor{gray!10} 3DSL~\cite{chen2021learning} & RGB+pose+sil.+3D & $\checkmark$ & 51.3 & \multicolumn{1}{c|}{-} & -  & - & 31.2 & \multicolumn{1}{c|}{14.8} & -      & -    \\
\rowcolor{gray!10} UCAD~\cite{yan2022weakening} & RGB+sil. & $\checkmark$ & 45.3 & \multicolumn{1}{c|}{-} & 96.5 & - & 32.5 & \multicolumn{1}{c|}{15.1} & 74.4 & 34.8 \\
\rowcolor{gray!10} CAL~\cite{gu2022clothes} & RGB & $\checkmark$ & 55.2 & \multicolumn{1}{c|}{55.8} & 100 & 99.8 & 40.1 & \multicolumn{1}{c|}{18.0} & 74.2 & 40.8 \\ 
\rowcolor{gray!10} DCR-ReID~\cite{cui2023dcr} & RGB+sil.+sketch & $\checkmark$ & 57.2 & 57.4 & 100 & 99.7 & 41.1 & 20.4 & 76.1 & 42.3 \\
\rowcolor{gray!10} AIM~\cite{yang2023good} & RGB & $\checkmark$ & 57.9 & \multicolumn{1}{c|}{58.3} & 100 & 99.9 & 40.6 & \multicolumn{1}{c|}{ 19.1} &  76.3 & 41.1 \\
\rowcolor{gray!10} CCFA$^{\ast}$~\cite{han2023clothing} & RGB & $\checkmark$ & 61.2 & \multicolumn{1}{c|}{58.4} & 99.6 & 98.7 & 45.3 & \multicolumn{1}{c|}{22.1} & 75.8 & 42.5 \\ \hline

HA-CNN~\cite{li2018harmonious} & RGB & $\scalebox{0.75}{\usym{2613}}$ & 21.8 & \multicolumn{1}{c|}{-} & 82.5 & - & 21.6 & \multicolumn{1}{c|}{9.3} & 60.2 & 26.7 \\
PCB~\cite{sun2018beyond} & RGB & $\scalebox{0.75}{\usym{2613}}$  & 41.8 & \multicolumn{1}{c|}{38.7} & 99.8   & 97.0 & 23.5 & \multicolumn{1}{c|}{10.0} & 65.1 & 30.6 \\
IANet~\cite{hou2019interaction} & RGB & $\scalebox{0.75}{\usym{2613}}$  & 46.3 & \multicolumn{1}{c|}{45.9} & 99.4 & 98.3 & 25.0 & \multicolumn{1}{c|}{12.6} & 63.7 & 31.0 \\
TransReID~\cite{he2021transreid} & RGB  & $\scalebox{0.75}{\usym{2613}}$  & 46.6  & \multicolumn{1}{c|}{44.8} & \bf 100 & \multicolumn{1}{c|}{99.0} & 34.4 & \multicolumn{1}{c|}{17.1}  & 70.4 & 37.0  \\
SPT+ASE~\cite{yang2019person} & sketch & $\scalebox{0.75}{\usym{2613}}$  & 34.4 & \multicolumn{1}{c|}{-} & 64.2  & -  & -  & \multicolumn{1}{c|}{-}  & -  & - \\
RCSANet$^{\ast}$~\cite{huang2021clothing} & RGB & $\scalebox{0.75}{\usym{2613}}$  & 50.2 & \multicolumn{1}{c|}{48.6} & \bf 100 & 97.2 & - & \multicolumn{1}{c|}{-} & - & -    \\
FSAM~\cite{hong2021fine} & RGB+pose+sil. & $\scalebox{0.75}{\usym{2613}}$  & 54.5 & \multicolumn{1}{c|}{-} & 98.8 & - & 38.5 & \multicolumn{1}{c|}{16.2} & 73.2 & 35.4 \\
GI-ReID$^{\ast}$~\cite{jin2022cloth} & RGB+sil. & $\scalebox{0.75}{\usym{2613}}$  & 33.3 & \multicolumn{1}{c|}{-} & 80.0  & -  & 23.7 & \multicolumn{1}{c|}{10.4} & 63.2   & 29.4 \\
CRE+BSGA~\cite{mu2022learning} & RGB+sil. & $\scalebox{0.75}{\usym{2613}}$  & 61.8 & \multicolumn{1}{c|}{58.7} & 99.6 & \multicolumn{1}{c|}{97.3} & - & \multicolumn{1}{c|}{-} & - & - \\
CAMC~\cite{wang2022co} & RGB+pose & $\scalebox{0.75}{\usym{2613}}$  & - & \multicolumn{1}{c|}{-} & - & \multicolumn{1}{c|}{-} & 36.0 & \multicolumn{1}{c|}{15.4} & 73.2 & 35.3 \\
\hline

FIRe$^{2}$ (Ours) & RGB & $\scalebox{0.75}{\usym{2613}}$  & \textbf{65.0} & \multicolumn{1}{c|}{\textbf{63.1}} & \textbf{100} & \bf 99.5 & \textbf{44.6} & \multicolumn{1}{c|}{\bf 19.1} & \bf 75.9 & \bf 39.9 \\ 
\Xhline{1pt}
\end{tabular}}
\vspace{-0.1in}
\end{table*}

\subsection{Evaluation Datasets}
To demonstrate the effectiveness of our proposed method, we evaluate it on three widely-used cloth-changing person Re-ID datasets, \textit{i.e.}, PRCC~\cite{yang2019person}, LTCC~\cite{qian2020long}, and Celeb-reID~\cite{huang2019beyond}, as well as two large-scale cloth-changing datasets, \textit{i.e.}, DeepChange~\cite{xu2023deepchange} and LaST~\cite{shu2021large}. 
Furthermore, we investigate the feasibility of our method for the cases without clothing changes, using two standard person Re-ID datasets in short-term scenarios, \textit{i.e.}, Market-1501~\cite{market1501} and MSMT17~\cite{wei2018person}. 

\noindent \textbf{(1) PRCC} contains 33,698 images of 221 identities captured from 3 indoor cameras. For each person, images from cameras A and B (C) are without (with) clothing changes. 
\textbf{LTCC} is an indoor cloth-changing person Re-ID dataset, which has 17,119 images of 152 identities captured from 12 camera views. LTCC is challenging as it contains diverse human poses, large changes of illumination, and large variations of occlusion.
\textbf{Celeb-reID} is acquired from the Internet using street snapshots of celebrities, and contains 34,186 images of 1,052 identities. Specifically, more than 70\% images of each person show different clothes on average. 

\noindent \textbf{(2) DeepChange} contains 178,407 images of 1,121 identities recorded by 17 outdoor varying resolution cameras operating in a real-world surveillance system.
\textbf{LaST} includes 10,862 identities with 228,156 images, captured from a broad range of countries, person ages, scenes, weathers, daytime and night, so it presents highly challenging and diverse.

\noindent \textbf{(3) Market-1501} is collected in Tsinghua University, including 32,668 images of 1,501 identities with 6 camera views. 
\textbf{MSMT17} is a large-scale benchmark captured in the morning, noon and afternoon on campus, containing 126,441 images from 4,101 identities captured by 15 cameras. 
Each pedestrian from both datasets always wears the same clothes.

\subsection{Experimental Setup} \label{subsec:exp_setup}

\noindent \textbf{Implementation details.}
Our method is implemented on the PyTorch framework. For fair comparison, following~\cite{wang2022co,gu2022clothes,yang2023good,han2023clothing}, we adopt ResNet-50~\cite{resnet} initialized by ImageNet~\cite{deng2009imagenet} as the backbone to extract features. Following~\cite{qian2020long,gu2022clothes,yan2022weakening,yang2023good}, the input images are resized to $384 \times 192$. For data augmentation, random horizontal flipping, padding, random cropping, and random erasing~\cite{zhong2020random} are used. Adam optimizer~\cite{kingma2014adam} with weight decay of $5 \times 10^{-4}$ is adopted for 80 epochs, with the warmup strategy that linearly increases the learning rate from $3.5 \times 10^{-6}$ to $3.5 \times 10^{-4}$ in the first 10 epochs, then decreases the learning rate by a factor of 10 every 20 epochs. We use $\mathcal{L}_{id}$ before learning rate decay and then introduce $\mathcal{L}_{tri}$, $\mathcal{L}_{attr}$, and $\mathcal{L}_{r}$ to optimize the model jointly. 
Hyper-parameters $\epsilon$, $\tau$, $P$ and $\lambda_{4}$ are set to 0.1, $1/16$, 2 and 0.3. The batch size is set to 64 for DeepChange and LaST, and 32 for other cloth-changing datasets. 
We use DBSCAN~\cite{ester1996density} for clustering with a minimum sample size of 1 set for each cluster. Setting the scanning radius to 0.4 mostly works well.

\noindent \textbf{Evaluation metrics.} 
For evaluation, we adopt standard metrics as in most person Re-ID literature, namely Cumulative Matching Characteristic (CMC) curves and mean Average Precision (mAP). To make a fair comparison with the existing research works, for LTCC and PRCC, we evaluate our method under both the standard setting and the cloth-changing setting. 
Specifically, for the standard setting of LTCC, there are both cloth-consistent and cloth-changing samples in the testing set. For the cloth-changing setting of both LTCC and PRCC, only cloth-changing samples are involved in the testing set.

\subsection{Comparison with State-of-the-Art Methods}
\label{subsec:comparison}
We compare our method with the baseline results using $\mathcal{L}_{id}$ (denoted by ``Baseline''), the strong baseline results using both $\mathcal{L}_{id}$ and $\mathcal{L}_{tri}$ (following~\cite{luo2019bag}, denoted by ``Baseline \textit{w/} Tri.''), and other state-of-the-art methods that have similar or the same experimental settings as ours for a fair comparison.

\noindent \textbf{Results on PRCC and LTCC.} 
As shown in Tab.~\ref{tab:ltcc}, we compare our proposed method with traditional person Re-ID methods and state-of-the-art methods especially designed for cloth-changing person Re-ID on PRCC and LTCC. Since appearance features are no longer useful, standard person Re-ID methods~\cite{li2018harmonious,sun2018beyond,hou2019interaction} achieve relatively inferior performance. TransReID~\cite{he2021transreid} exploits the stronger capacity of Transformer for Re-ID. Though its performance is even comparable to some cloth-changing baselines, our method achieves much more significant advantages on both datasets, showing the superiority of our proposed modules.

Among cloth-changing person Re-ID methods, some of them \cite{yang2019person,hong2021fine,jin2022cloth,mu2022learning,wang2022co} resort to other modalities for help (\textit{e.g.}, contour sketches and human poses) to avoid the interference of clothes. However, our method only takes RGB modality as input and outperforms them by a large margin. Others \cite{qian2020long,xu2021adversarial,chen2021learning,yan2022weakening,gu2022clothes,yang2023good,han2023clothing,cui2023dcr} depend on manually annotated clothing labels to be aware of or eliminate the influence of clothes. Differently, we adequately mine more discriminative fine-grained identity information without any annotations. Our FIRe$^{2}$ achieves state-of-the-art results of 65.0/63.1\% Rank-1/mAP on PRCC, and gets comparable performance of 44.6/19.1\% on LTCC. More importantly, without the requirement of clothing labels, our method can be much more easily applied to other scenarios or datasets.

\begin{table}[t]
\centering
\caption{\label{tab:celeb}\textbf{Comparisons with state-of-the-art methods on Celeb-reID.} ``pose'' and ``gray'' represent human poses and gray images, respectively. Methods marked with ``$\star$'' adopt DenseNet-121~\cite{huang2017densely} as the backbone.
}
\vspace{-0.05in}
\setlength{\tabcolsep}{2mm}{
\begin{tabular}{l|c|ccc}
\Xhline{1pt}
\multicolumn{1}{c|}{\textbf{Methods}} & \textbf{Modality} & \textbf{Rank-1} & \textbf{Rank-5} & \textbf{mAP}  \\ 
\hline \hline
Part-Bilinear~\cite{suh2018part} & RGB & 19.4   & 40.6   & 6.4  \\
PCB~\cite{sun2018beyond} & RGB & 37.1   & 57.0   & 8.2  \\
MGN~\cite{wang2018learning} & RGB & 49.0   & 64.9   & 10.8 \\ 

ReIDCaps~\cite{huang2019beyond} & RGB & 51.2   & 65.4   & 9.8  \\
CESD~\cite{qian2020long} & RGB+pose & 50.9   & 66.3   & 9.8  \\
AFD-Net~\cite{xu2021adversarial} & RGB+gray & 52.1   & 66.1   & 10.6 \\
IS-GAN$_{KL}$~\cite{eom2021disentangled} & RGB & 54.5   & -      & 12.8 \\ 
RCSANet$^{\star}$~\cite{huang2021clothing} & RGB & 55.6   & -      & 11.9 \\ 
SirNet$^{\star}$~\cite{yang2022sampling} & RGB+gray & 56.0 & 70.3 & 14.2 \\ 
CAMC~\cite{wang2022co} & RGB+pose & 57.5 & 71.5 & 12.3 \\ 
\hline
Baseline & RGB & 54.3     & 67.4      & 10.5     \\
Baseline \textit{w/} Tri. & RGB & 56.9     & 70.3      & 11.6     \\
FIRe$^{2}$ (Ours)  & RGB  & \textbf{64.0}   & \textbf{78.8}   & \textbf{18.2}  \\ \Xhline{1pt}
\end{tabular}}
\vspace{-0.15in}
\end{table}

\noindent \textbf{Results on Celeb-reID.} 
We also evaluate our method on the Celeb-reID dataset. It is a more large and challenging dataset, where images are captured from uncontrolled street snapshots. It thus has dramatic changes in the attributes of each individual and does not contain any clothing annotation. As shown in Tab.~\ref{tab:celeb}, all advanced methods achieve relatively poor performance, even those using stronger backbone models~\cite{huang2021clothing,yang2022sampling} and specially designed for the cloth-changing task~\cite{qian2020long,xu2021adversarial,huang2021clothing,eom2021disentangled,yang2022sampling,wang2022co}. In addition, competitors like CAL~\cite{gu2022clothes}, AIM~\cite{yang2023good} and CCFA~\cite{han2023clothing} fail to report results on this dataset due to the lack of clothing labels. However, without relying on extra modalities or clothing labels, FIRe$^{2}$ achieves remarkable performance of 64.0\% and 18.2\% on Rank-1 and mAP metrics, thanks to the FFM module.

\begin{table}[t]
\centering
\caption{\label{tab:deepchange}\textbf{Comparison of our method with state-of-the-art methods on the DeepChange dataset.} CAL uses the collection date of each image as pseudo clothing labels for training.
}
\vspace{-0.05in}
\setlength{\tabcolsep}{4mm}{
\begin{tabular}{l|c|cc}
    \Xhline{1pt}
    \multicolumn{1}{c|}{\multirow{2}{*}{\textbf{Methods}}} & \textbf{Cloth-} & \multirow{2}{*}{\textbf{Rank-1}} & \multirow{2}{*}{\textbf{mAP}} \\
     & \textbf{Labels} & &  \\
    \hline \hline
    OSNet~\cite{zhou2019omni} & $\scalebox{0.75}{\usym{2613}}$ & 39.7          & 10.3          \\
    ReIDCaps~\cite{huang2019beyond} & $\scalebox{0.75}{\usym{2613}}$ &  39.5          & 11.3          \\
    ViT-B/16~\cite{xu2023deepchange} & $\scalebox{0.75}{\usym{2613}}$ & 49.8 & 15.0 \\
    SCNet~\cite{guo2023semantic} & $\scalebox{0.75}{\usym{2613}}$ & 53.5 & 18.7 \\
    \rowcolor{gray!10} CAL~\cite{gu2022clothes} & $\checkmark$ & 54.0          & 19.0 \\ 
    \hline
    Baseline & $\scalebox{0.75}{\usym{2613}}$ &  51.6         & 16.0          \\
    Baseline \textit{w/} Tri. & $\scalebox{0.75}{\usym{2613}}$ &  52.5         & 16.1          \\
    FIRe$^{2}$ (Ours) & $\scalebox{0.75}{\usym{2613}}$ & \textbf{57.9} & \textbf{20.0}  \\ \Xhline{1pt}
\end{tabular}}
\vspace{-0.1in}
\end{table}

\begin{table}[t]
\centering
\caption{\label{tab:last}\textbf{Comparisons with state-of-the-art methods on the LaST dataset.} ``\textit{w/} Tri.'' means additionally applying triplet loss.
}
\vspace{-0.05in}
\setlength{\tabcolsep}{3.8mm}{
\begin{tabular}{l|c|cc}
    \Xhline{1pt}
    \multicolumn{1}{c|}{\multirow{2}{*}{\textbf{Methods}}} & \textbf{Cloth-} & \multirow{2}{*}{\textbf{Rank-1}} & \multirow{2}{*}{\textbf{mAP}} \\
     & \textbf{Labels} & &  \\
    \hline \hline
    OSNet~\cite{zhou2019omni} & $\scalebox{0.75}{\usym{2613}}$  & 63.8          & 20.9          \\
    mAPLoss~\cite{shu2021large} & $\scalebox{0.75}{\usym{2613}}$  & 69.9          & 27.6          \\
    \rowcolor{gray!10} CAL~\cite{gu2022clothes} \textit{w/} Tri. & $\checkmark$  & 73.7          & 28.8 \\ \hline
    Baseline & $\scalebox{0.75}{\usym{2613}}$  & 70.5      &  26.7     \\
    Baseline \textit{w/} Tri. & $\scalebox{0.75}{\usym{2613}}$  & 71.7      &  28.2    \\
    FIRe$^{2}$ (Ours) & $\scalebox{0.75}{\usym{2613}}$  & \textbf{75.0} & \textbf{32.2}  \\ \Xhline{1pt}
\end{tabular}}
\vspace{-0.15in}
\end{table}

\noindent \textbf{Results on DeepChange and LaST.} 
Furthermore, we evaluate our method on the DeepChange and LaST datasets. The scale of both datasets is much larger than previous benchmarks, especially the number of gallery samples for testing, reflecting the complexity and difficulty in real-world scenarios. 
However, most state-of-the-art methods do not report results on these datasets. Without loss of generality, we compare with several strong baselines and competitors using their official code.

Results on the DeepChange dataset are shown in Tab.~\ref{tab:deepchange}. Our proposed FIRe$^{2}$ outperforms the ResNet-50~\cite{resnet} baseline and ViT-B/16~\cite{dosovitskiy2020image,xu2023deepchange} substantially, and surpasses the state-of-the-art method CAL~\cite{gu2022clothes} with 2.6\% mAP. 
SCNet~\cite{guo2023semantic} additionally uses the prior knowledge of human parsing for feature enhancement or suppression. We also outperform it, which shows the effectiveness and better generalization ability of our method on the large-scale dataset. Table~\ref{tab:last} demonstrates the results on the LaST dataset. Although clothing labels are available in LaST, our method does not rely on any auxiliary information, and achieves the best performance. 
We observe that CAL~\cite{gu2022clothes}, a competitor that uses annotated clothing labels to optimize the model in an adversarial way, is still inferior to our FIRe$^{2}$. Such results strongly indicate the efficacy of our method, especially the design of fine-grained feature learning and attribute recomposition module for cloth-changing person Re-ID. It is worth mentioning that our proposed method has no assumption on the number of images. Even with a minimal number of images per identity of only 5 on LaST, FIRe$^{2}$ still works better than other competitors, showing its robustness.

\begin{table}[t]
\centering
\caption{\label{tab:market-msmt17}\textbf{Comparisons with state-of-the-art methods on the Market-1501 and MSMT17 datasets.}
}
\vspace{-0.05in}
\setlength{\tabcolsep}{2.7mm}{
\begin{tabular}{l|cc|cc}
\Xhline{1pt}
\multicolumn{1}{c|}{\multirow{2}{*}{\textbf{Methods}}} & \multicolumn{2}{c|}{\textbf{Market-1501}} & \multicolumn{2}{c}{\textbf{MSMT17}} \\ \cline{2-5} 
                                & Rank-1 & mAP  & Rank-1 & mAP  \\ \hline \hline
PCB~\cite{sun2018beyond}        & 93.8   & 81.6 & 68.2   & 40.4  \\
IANet~\cite{hou2019interaction} & 94.4   & 83.1 & 75.5   & 46.8  \\
OSNet~\cite{zhou2019omni}      & 94.8   & 84.9 & 78.7   & 52.9  \\
JDGL~\cite{zheng2019joint}      & 94.8   & 86.0 & 77.2   & 52.3  \\
CircleLoss~\cite{sun2020circle} & 94.2   & 84.9 & 76.3   & 50.2  \\ \hline
Baseline                        & 91.8   & 80.6 & 70.8   & 45.6  \\
Basline \textit{w/} Tri.        & 92.5   & 82.1 & 73.7   & 48.6  \\
FIRe$^{2}$ (Ours)               & \textbf{95.4}   & \textbf{87.7} & \textbf{79.7}   & \textbf{56.2}  \\ \Xhline{1pt}
\end{tabular}}
\vspace{-0.1in}
\end{table}

\noindent \textbf{Results on Market-1501 and MSMT17.}
Lastly, we evaluate our method on conventional benchmarks to verify its generalization ability in short-term scenarios. Although there is no change of clothes in both datasets, there are still various variations (\textit{e.g.}, pose, viewpoint and occlusion). We can effectively mine fine-grained attributes leveraging our FFM module and further augment representations with them through our FAR module. As shown in Tab.~\ref{tab:market-msmt17}, our method can greatly outperform the baseline models and achieve comparable performance with state-of-the-art competitors. Such remarkable results demonstrate that the robust fine-grained identity information can also effectively promote the recognition ability of the model on standard person Re-ID benchmarks.

\subsection{Ablation Studies}
\label{subsec:ablation}
In this section, we perform comprehensive ablation studies on PRCC and LTCC. We discuss in detail the efficacy of the proposed modules and the effect of hyper-parameters.

\begin{figure*}[t]
  \centering
  \includegraphics[width=0.8\linewidth]{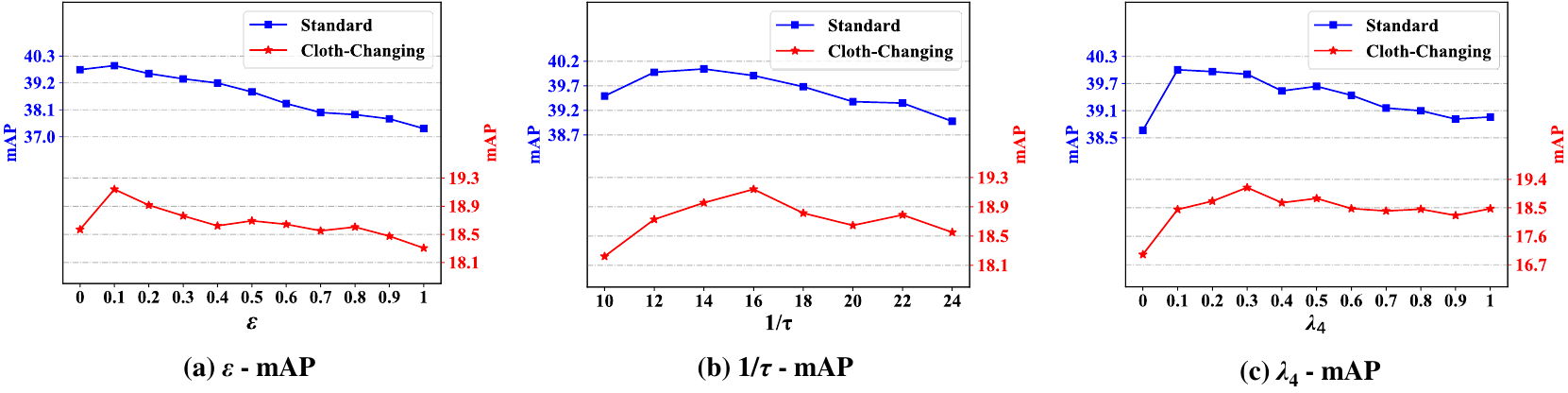}
  \vspace{-0.1in}
  \caption{\textbf{Ablation studies of hyper-parameters.} We report mAP results of our method with different values of (a) $\epsilon$, (b) $1 / \tau$, (c) $\lambda_{4}$ on the LTCC dataset. ``Standard’' and ``Cloth-Changing'' mean the standard and cloth-changing settings, respectively.}
  \label{fig:3hyper_params}
  \vspace{-0.15in}
\end{figure*}

\begin{table}[t]
\centering
\caption{\label{tab:ablation}\textbf{Ablation studies of our method on PRCC and LTCC under the cloth-changing setting.} ``Ours \textit{w/} Cloth.'' means replacing the fine-grained pseudo labels used in our method with the ground-truth clothing labels. ``Ours \textit{w/} mixup'' means replacing our proposed FAR with mixup~\cite{zhang2018mixup}.}
\vspace{-0.05in}
\setlength{\tabcolsep}{2.6mm}{
\begin{tabular}{l|cc|cc}
\Xhline{1pt}
\multicolumn{1}{c|}{\multirow{2}{*}{\bf Methods}} & \multicolumn{2}{c|}{\bf PRCC}     & \multicolumn{2}{c}{\bf LTCC}      \\ \cline{2-5}
                         & Rank-1        & mAP           & Rank-1        & mAP           \\ \hline \hline
Baseline                 & 40.2          & 36.2          & 28.8          & 11.5          \\
Baseline \textit{w/} Tri. & 48.3          & 46.6          & 33.2          & 14.2          \\
Ours \textit{w/} Cloth.  & 61.5          & 61.1          & 40.3 & 18.4          \\
Ours \textit{w/o} $\mathcal{L}_{attr}$ & 63.6        & 62.2          & 41.6 & 18.5 \\
Ours \textit{w/o} FAR    & 62.1          & 61.5          & 40.3 & 17.5 \\
\hline
Ours \textit{w/} mixup~\cite{zhang2018mixup} & 63.5 & 61.4 & 40.1 & 17.2 \\
FAR within ID & 63.8 & 62.8 & 40.8 & 17.4 \\
FAR between IDs & 62.7 & 62.6 & 43.1 & 18.7 \\
\hline
FIRe$^{2}$ (Ours)        & \textbf{65.0} & \textbf{63.1} & \textbf{44.6} & \textbf{19.1} \\ \Xhline{1pt}
\end{tabular}}
\vspace{-0.15in}
\end{table}

\noindent \textbf{Effectiveness of fine-grained feature mining module.}
To show the rationality and effectiveness of the proposed label acquisition strategy, we replace the fine-grained pseudo labels used in our method with the ground-truth clothing labels provided in the datasets. As ``Ours \textit{w/} Cloth.'' shown in Tab.~\ref{tab:ablation}, it achieves competitive results with state-of-the-art methods~\cite{gu2022clothes,cui2023dcr,yang2023good}, showing the good generalization ability of our model, since clothing can be treated as one kind of fine-grained attribute. Moreover, our proposed FIRe$^2$ further beats ``Ours \textit{w/} Cloth.'' by a significant margin of 3.5\%/2.0\% and 4.3\%/0.7\% Rank-1/mAP on PRCC and LTCC, respectively. It clearly indicates that using clothing labels is not the upper bound of our model. In contrast, our proposed FFM module can produce more fine-grained attributes than just clothes, embracing more learning benefits.
In addition, the inferior results of ``Ours \textit{w/o} $\mathcal{L}_{attr}$'' compared with ``Ours'' demonstrate the effectiveness of our proposed attribute-aware classification loss. All in all, without extra labor costs for fine-grained annotations, our proposed FFM module effectively learns fine-grained information, facilitating the discriminative representations of cloth-changing person images.

\noindent \textbf{Efficacy of fine-grained attribute recomposition module.}
As shown in Tab.~\ref{tab:ablation}, with our proposed FAR module, which recomposes attributes based on fine-grained pseudo labels, the performance is greatly improved. Under the cloth-changing setting, the FAR module can bring 2.9\%/1.6\% Rank-1/mAP improvement on PRCC, and 4.3\%/1.6\% on LTCC. 
We also try adopting mixup~\cite{zhang2018mixup} as an augmentation manner to replace our FAR module. Results in Tab.~\ref{tab:ablation} show that ``Ours \textit{w/} mixup'' is slightly inferior to our model that excludes the attribute recomposition module (\textit{i.e.}, ``Ours \textit{w/o} FAR''). One possible reason is that the mixup approach mixes attributes from different people while also mixing their identity labels. It may contradict the motivation of our fine-grained attribute learning loss designed in the FAR module, thus leading to sub-optimal performance. The results further validate the efficacy and superiority of FAR in robust identity-related feature learning.

To show the effectiveness of our design that recomposes fine-grained attributes between either the same or different identities, we additionally conduct two variants of performing FAR only within the same identity (``FAR within ID'') and solely between different identities (``FAR between IDs''). As shown in Tab.~\ref{tab:ablation}, both variants achieve better results than ``Ours \textit{w/o} FAR'', but they are inferior to our full model by a clear margin. Interestingly, ``FAR between IDs'' has significant improvements on the LTCC dataset, since LTCC is collected in a more complicated scenario than PRCC and each person has multiple outfits. The results strongly support our motivation and demonstrate the superiority of our design in FAR.

\noindent \textbf{Influence of the hyper-parameter $\epsilon$ in Eq.~\ref{eq:w-loss}.}
We try different values of $\epsilon$ in Fig.~\ref{fig:3hyper_params}~(a) and have the following observations.
\textbf{(1)} When $\epsilon = 0$, the attribute-aware classifier would simply classify each feature to its corresponding fine-grained attribute class, overlooking the identity supervision and pushing different clusters with the same identity away. It thus leads to sub-optimal results.
\textbf{(2)} When $\epsilon$ is set to a small value (\textit{e.g.}, 0.1$\sim$0.3), the distances between clusters with the same identity are kept smaller than those with different identities. It shows a clear performance improvement, indicating the efficacy of our fine-grained representation learning.
\textbf{(3)} With $\epsilon$ continuing to increase, the performance dramatically decreases. The model receives confused large supervision signals from multiple fine-grained classes, distracting the learning of discriminative features. Even though, the results are still competitive with other baselines. Overall, we set $\epsilon$ to 0.1 for all experiments.

\noindent \textbf{Influence of the temperature factor $\tau$ in Eq.~\ref{eq:attr-loss}.}
As shown in Fig.~\ref{fig:3hyper_params}~(b), results are relatively stable for different values of $\tau$ in attribute-aware classification loss. The best results under the cloth-changing setting on LTCC are achieved when $1 / \tau = 16$, and we simply set it as default for all experiments.

\noindent \textbf{Influence of the coefficient $\lambda_{4}$ in Eq.~\ref{eq:overall_loss}.}
In Fig.~\ref{fig:3hyper_params}~(c), we further investigate different values of $\lambda_{4}$. We observe that
\textbf{(1)} When $\lambda_{4}$ is too small, we cannot give full play to the facilitation of the FAR module, resulting in inferior performance.
\textbf{(2)} When $\lambda_{4}$ is set to a large value (\textit{e.g.}, 0.6 $\sim$ 1), the results are still sub-optimal, since the model may be misguided by hard attribute-recomposed features and the domain gap between the real and synthesized samples. Meanwhile, a too large value of $\lambda_{4}$ is not friendly to the standard setting, where each person would always wear the same clothes and there is less attribute variation.
\textbf{(3)} With a proper loss contribution, the model can benefit from augmented attribute-recomposed features to be more robust to attribute variation. Here, we set $\lambda_{4} = 0.3$ throughout the experiments.

\begin{figure}[t]
  \centering
  \includegraphics[width=0.94\linewidth]{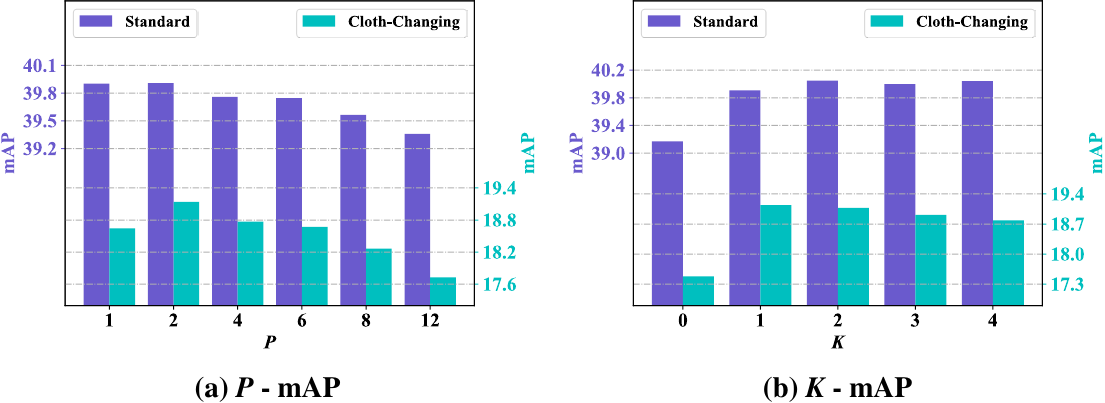}
  \vspace{-0.05in}
  \caption{\textbf{Ablation studies of fine-grained attribute recomposition.} We report mAP results with different values of (a) body parts $P$, and (b) attribute recompostion times $K$ on LTCC. ``Standard’' and ``Cloth-Changing'' mean the standard and cloth-changing settings, respectively.
  }
  \label{fig:PK_hyper_params}
  \vspace{-0.15in}
\end{figure}

\noindent \textbf{Influence of the number of body parts $P$.}
Considering the fine-grained attribute inconsistency across the spatial, we equally partition each feature into $P$ parts horizontally and perform attribute recomposition within each part. As shown in Fig.~\ref{fig:PK_hyper_params}~(a), when $P$ is set to 1, \textit{i.e.}, there is no part partition, the good performance shows the effectiveness of our proposed FAR module. When we partition features into too many parts, each part will be small. We cannot get useful attributes with small parts, since attributes are presented as statistical mean and standard deviation. Therefore, large $P$ brings fewer gains. Considering the efficiency and performance, we set $P$ to 2.

\noindent \textbf{Influence of attribute recomposition times $K$.}
With the help of efficient recomposition as described in Sec.~\ref{subsec:FAR}, we are able to adopt an intuitive trick for augmentation to generate $K \ge 2$ attribute-recomposed features by sampling multiple fine-grained part attributes within a batch and performing Eq.~\ref{eq:adain}. We explore the influence of the number of attribute recomposition times $K$ in Fig.~\ref{fig:PK_hyper_params}~(b). With the increase of $K$, the model achieves similar performance. We argue that attribute recomposition serves as data augmentation in the latent space, so it is enough to guide the model to learn robust fine-grained identity features with $K = 1$. Considering the efficiency and performance, we simply set $K$ to~1.

\begin{figure}[t]
\centering
  \includegraphics[width=0.91\linewidth]{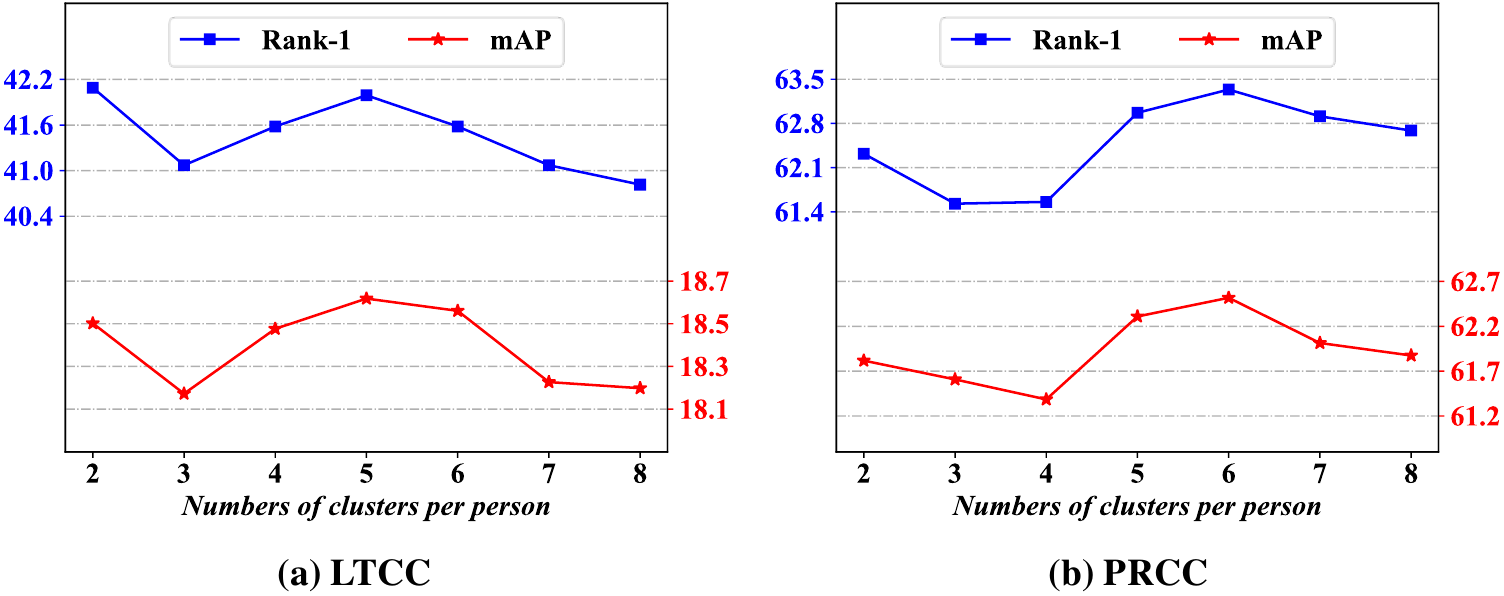}
  \vspace{-0.05in}
  \caption{\textbf{Influence of setting a fixed number of clusters per person.} The results correspond to the cloth-changing setting of (a) LTCC and (b) PRCC.}
  \label{fig:fix_cluster_num}
  \vspace{-0.15in}
\end{figure}

\noindent \textbf{More discussions on clustering hyper-parameters.}
As stated in Sec.~\ref{sec:intro}, it is ambiguous and impractical to manually define clothing numbers for different IDs. Therefore, we adopt the DBSCAN~\cite{ester1996density} algorithm for fine-grained feature mining efficiently and adaptively. One of the important factors in clustering is the number of clusters. Firstly, we adjust the number of clusters indirectly by tuning the scanning radius within the range of $0.4\pm0.2$. We find that mAP only fluctuates within 0.5\% on both LTCC and PRCC, all exceeding baselines significantly. 
Furthermore, we conduct experiments by manually fixing different clustering numbers for each person. Results in Fig.~\ref{fig:fix_cluster_num} show their inferiority to our full model equipped with the FFM module to dynamically unearth different kinds of attributes of each person. Besides, the optimal choice varies among datasets and needs to be cherry-picked. It indicates the effectiveness and generalization ability of our FFM module.

\noindent \textbf{More comparisons on time cost.}
Last but not least, we discuss the time complexity of our method in terms of FLOPs, training and inference time, to show its potential in the real-world deployment and application. More concretely, the FLOPs of FIRe$^{2}$ is $9.2$G, which is significantly lighter than $18.5$G for AIM~\cite{yang2023good} and $19.7$G for CAMC~\cite{wang2022co}. Both competitors get inferior or comparable performance to ours in Tab.~\ref{tab:ltcc}. 
As for the time cost, it takes about 2.4 and 4.7 hours to train our FIRe$^{2}$ on the LTCC and PRCC datasets, respectively. By contrast, the training time is about 4.3 and 7.8 hours for AIM~\cite{yang2023good}; and 5.3 and 10 hours for CAMC~\cite{wang2022co} on the same device. 
More importantly, thanks to our design, the proposed modules are only involved during training to facilitate fine-grained representation learning, and could be omitted during inference to avoid more storage or computing overload. Based on the above analysis, FIRe$^{2}$ shows obvious superiority over other competitors in both efficacy and efficiency.

\subsection{Visualizations}
\label{sebsec:visualizations}

\begin{figure}[t]
  \centering
  \includegraphics[width=0.91\linewidth]{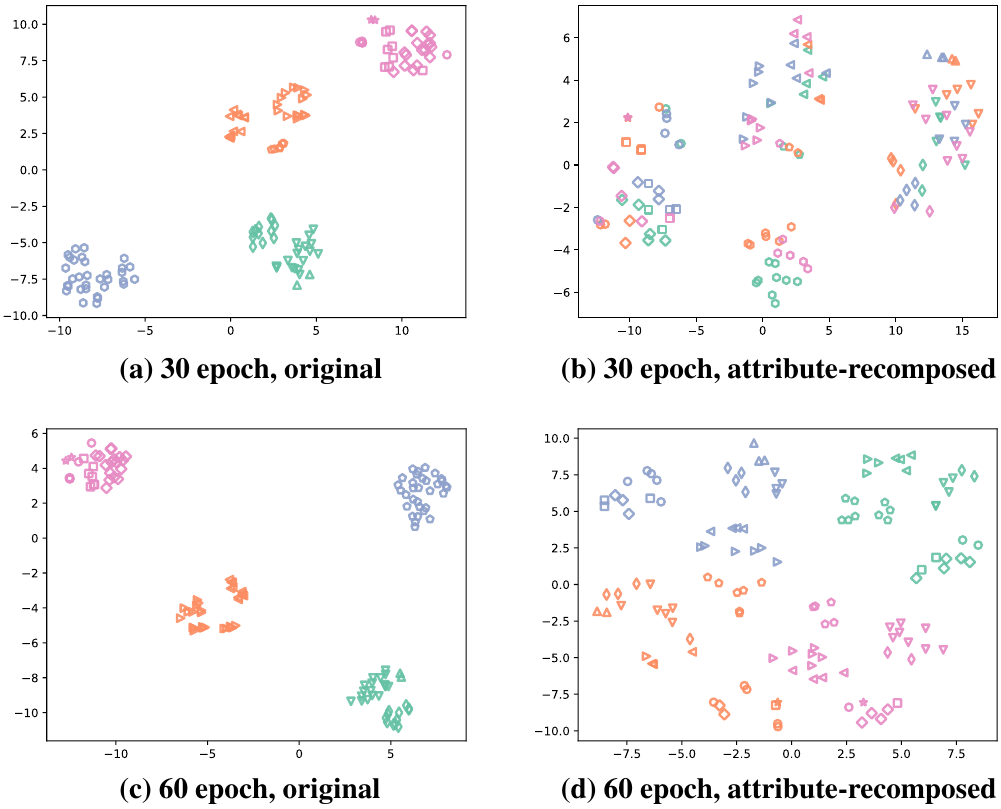}
  \vspace{-0.05in}
  \caption{\textbf{t-SNE visualization of the distribution of features before and after our proposed FAR module at different epochs.} 
  Each color represents an identity, and symbols denote different fine-grained pseudo labels. For simplicity, symbols in (b) and (d) refer to recomposed attributes of the upper body. Best viewed in color and zoomed in.}
  \label{fig:attr-recomposed-feat}
  \vspace{-0.15in}
\end{figure}

To better understand the mechanism of our method and the efficacy of each designed module, we further provide qualitative results of features, clustering and retrieval.

\noindent \textbf{Visualizations of recomposed features.}
To investigate the efficacy of the FAR module, which recomposes attributes in the feature space, we use t-SNE~\cite{van2008visualizing} to visualize the distribution of features before and after it at different epochs. 
\textbf{(1)} From Fig.~\ref{fig:attr-recomposed-feat}~(a) to (c), images with the same fine-grained pseudo label are gradually gathered. Besides, different identities are further pushed away. 
\textbf{(2)} More interestingly, the number of fine-grained pseudo labels of yellow samples is dynamically updated during fine-grained learning.
\textbf{(3)} As shown in Fig.~\ref{fig:attr-recomposed-feat}~(b), at the early training stage, features are sensitive to attribute variation, thereby the clustering results are dominated by recomposed attributes, ignoring their own identities. These recomposed features as hard negative samples force the model to learn more discriminative and identity-relevant representations.
\textbf{(4)} When the training approaches convergence, the model can distinguish the identity of image features even after fine-grained attribute recomposition. In this case, FAR can generate more hard positive samples, expanding the distribution range of each identity in the feature space, to improve the model's robustness, as demonstrated in Fig.~\ref{fig:attr-recomposed-feat}~(d). Eventually, even with attribute variation, all samples are distributed according to their identities.

\begin{figure}[t]
  \centering
  \includegraphics[width=0.94\linewidth]{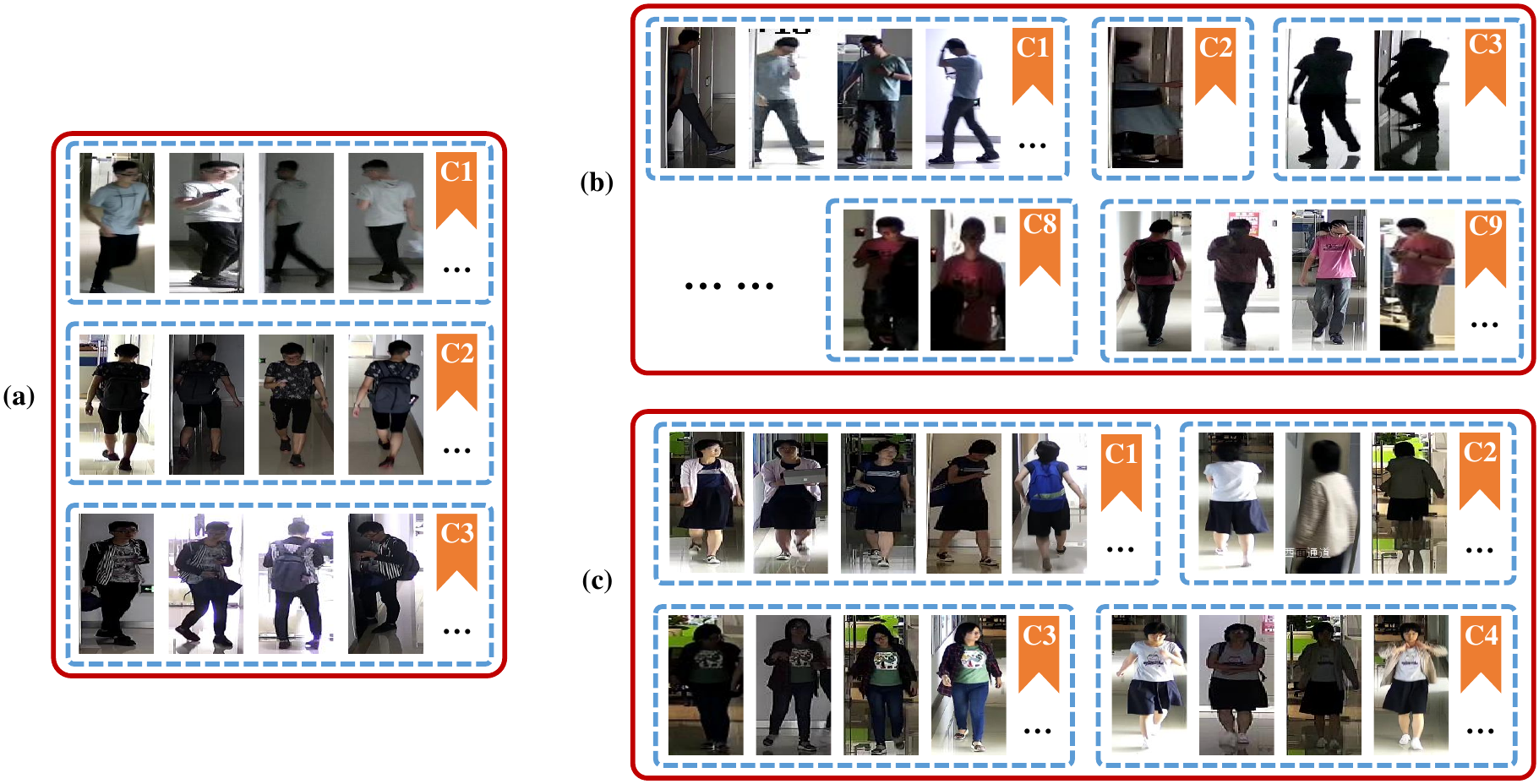}
  \vspace{-0.05in}
  \caption{\textbf{Clustering results of images from the same person.} We randomly select three identities from LTCC in (a)$\sim$(c). Images with the same identity are clustered with our FFM module.
  Best viewed in color and zoomed in.
  }
  \label{fig:more-cluster-imgs}
  \vspace{-0.15in}
\end{figure}

\noindent \textbf{Visualizations of clustering results.}
To further analyze how images with the same identity label are assigned to different clusters, we show more clustering results of images from the same person in Fig.~\ref{fig:more-cluster-imgs}. We make conclusions below based on observations.
\textbf{\textit{(1) Fine-grained pseudo labels are more robust against manual annotations to some extent.}} 
Manual annotations may inevitably involve mistakes. For example, in (b), the image in C2 is originally annotated as the same clothing label as that of images in C1 and C3, but obviously, it has different clothes and even is not the same identity as those in (b). 
Additionally, in (c), the first two images in C3 are originally labeled as different clothing from others in C3, but they have the same clothing. With FFM and our proposed attribute-aware fine-grained learning, we can effectively prevent the model from being misled by wrong labels.
\textbf{\textit{(2) Fine-grained pseudo labels help learn fine-grained information other than clothing.}}
As stated in Sec.~\ref{subsec:FFM}, our clustering is performed on the features of all images of each person separately. On the one hand, the number of clusters is mostly no less than the number of clothes per person.
On the other hand, the variable factors in person images are not only clothing, but also posture, occlusion, \textit{etc}. These variants naturally affect the image feature distribution, which is captured by different clusters after clustering, resulting in multiple fine-grained pseudo labels. As illustrated in C9 of (b) and C1 of (c), person images with different appearance features (\textit{e.g.}, knapsack or illumination) are clustered together due to the similar posture. 
One special case occurred in (a) that images are aggregated according to clothing because of the limited number of images and variations. It is reasonable, but more commonly, our FFM module can mine more diverse fine-grained information than clothing to facilitate fine-grained representation learning.
\textbf{\textit{(3) Fine-grained pseudo labels can capture rich implicit factors from images.}} 
Pedestrians even wearing the same clothes could be assigned into different clusters due to other fine-grained attributes. For example, images in C1 and C3 of (b) are originally annotated with the same clothing label, but are assigned two fine-grained pseudo labels because of different illumination. Similar phenomena can be found in C8 and C9 of (b), and C2 and C4 of (c), affected by viewpoint and poses respectively.
In sum, our proposed FFM module can effectively discover the variable factors in person images without any annotations.

\begin{figure}[t]
  \centering
  \includegraphics[width=0.92\linewidth]{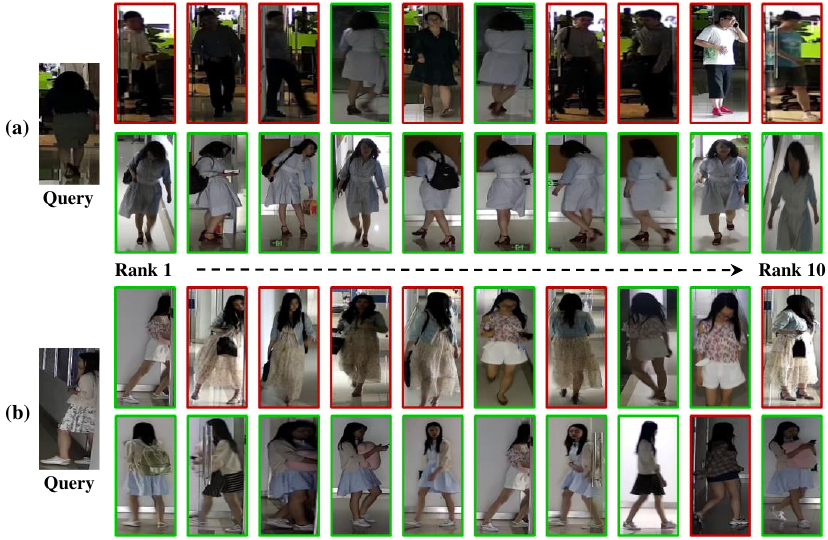}
  \vspace{-0.05in}
  \caption{\textbf{Visualization of top-10 retrieval results on LTCC.}
  For each query image, the first and the second rows are the ordered matching results obtained by using the ResNet-50 baseline and FIRe$^{2}$, respectively. Images with green and red borders indicate correct and wrong matching results, respectively.
  \label{fig:retrieval}}
\vspace{-0.15in}
\end{figure}

\noindent \textbf{Visualizations of retrieval results.}
To intuitively demonstrate the effect of our method for cloth-changing person Re-ID, we compare the retrieval results with the baseline model in Fig.~\ref{fig:retrieval}. FIRe$^{2}$ is more robust and can better recognize the same person with various clothes. For example, the baseline model is unable to identify pedestrians correctly with the interference of similar visual appearances in the first row results of (a) and (b). However, those pedestrian images with changing clothes are correctly retrieved by our method. We also observe some failure cases of our method, where one female wearing different clothes but with a similar figure is wrongly retrieved. Overall, benefiting from fine-grained and robust identity-relevant representations, our method can more effectively identify pedestrians in cloth-changing scenarios.

\section{Social Impact and Limitations}
\noindent \textbf{Broader Impact.}
In reality, person Re-ID systems typically use surveillance data, which may cause privacy problems due to unauthorized or illegal activities. Our proposed method captures fine-grained attributes and recomposes new features based on learned statistics of the training dataset. It may reflect biases in those data or generate unsafe content. Furthermore, governments and officials must take action to govern the use of person Re-ID data and technology, and researchers should be aware of negative societal impacts and avoid using datasets with ethical concerns. It is worth mentioning that all datasets used in our paper are publicly available.

\noindent \textbf{Limitations.} 
In our proposed framework, the clustering operation may inevitably bring additional memory consumption and time costs during training, and the quality of clustering may affect the final results. As an exploration attempt, we try to replace DBSCAN with SpCL~\cite{ge2020self}, an advanced clustering method that constrains the reliability and quality of each cluster. It achieves about 0.2\% gains of mAP on PRCC and LTCC datasets. The improvement, though marginal, is indeed inspiring. In the future, we will explore more advanced attribute clustering and recomposition methods, which are expected to achieve greater improvement.

\section{Conclusion}
In this paper, we propose a novel FIne-grained Representation and Recomposition (FIRe$^{2}$) framework for cloth-changing person Re-ID. Specifically, we first propose a Fine-grained Feature Mining module to obtain fine-grained pseudo labels and mine fine-grained attributes of each person through clustering. An attribute-aware classification loss is introduced to leverage pseudo labels for fine-grained learning. Additionally. we introduce a Fine-grained Attribute Recomposition module, which takes advantage of the explored fine-grained attributes to enrich feature representations by recomposing attributes between images.
FIRe$^{2}$ achieves state-of-the-art performance on several benchmarks. We hope it can inspire more research to focus on fine-grained learning for cloth-changing person Re-ID, without relying on expensive and impractical clothing annotations or other inflexible auxiliary information.

\bibliographystyle{IEEEtran}
\bibliography{egbib}

\end{document}